\documentclass[10pt,twocolumn,letterpaper]{article}

\usepackage{iccv}
\usepackage{times}
\usepackage{epsfig}
\usepackage{graphicx}
\usepackage{amsmath}
\usepackage{amssymb}

\usepackage{epsfig}
\usepackage{graphicx}
\usepackage{amsmath}
\usepackage{amssymb}
\usepackage{bm}
\usepackage{xcolor}
\usepackage{tabu}
\usepackage{color,array}
\usepackage{multirow,makecell}

\usepackage{caption}
\usepackage{subcaption}

\iccvfinalcopy 

\def\iccvPaperID{0008} 

\ificcvfinal\fi

\begin{document}

\title{A Semi-Supervised Maximum Margin Metric Learning Approach for Small Scale Person Re-identification}

\author{T M Feroz Ali\\
Indian Institute of Technology Bombay\\
{\tt\small ferozalitm@ee.iitb.ac.in}
\and
Subhasis Chaudhuri\\
Indian Institute of Technology Bombay\\
{\tt\small sc@ee.iitb.ac.in}
}

\maketitle
\ificcvfinal\thispagestyle{empty}\fi

\begin{abstract}
In video surveillance, person re-identification is the task of searching person images in non-overlapping cameras. Though supervised methods for person re-identification have attained impressive performance, obtaining large scale cross-view labeled training data is very expensive. However, unlabelled data is available in abundance.  In this paper, we propose a semi-supervised metric learning approach that can utilize information in unlabelled data with the help of a few labelled training samples.  We also address the small sample size problem that inherently occurs due to the few labeled training data. Our method learns a discriminative space where within class samples collapse to singular points, achieving the least within class variance, and then use a maximum margin criterion over a high dimensional kernel space to maximally separate the distinct class samples. A maximum margin criterion with two levels of high dimensional mappings to kernel space is used to obtain better cross-view discrimination of the identities. Cross-view affinity learning with reciprocal nearest neighbor constraints is used to mine new pseudo-classes from the unlabelled data and update the distance metric iteratively. We attain state-of-the-art performance on  four challenging datasets with a large margin.
\end{abstract}

\section{Introduction}
\begin{figure}[t!]
\begin{center}
   \includegraphics[width=0.95\linewidth]{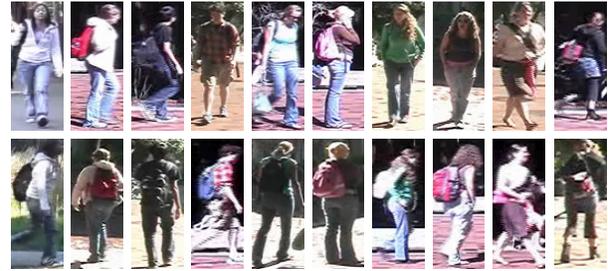}
\end{center}
   \caption{Challenges in person re-identification: large variations in viewpoint and illumination across camera views (each row represents different camera view).}
\label{fig:ReIDChallenge}
\end{figure}

Person re-identification (re-ID) is the task of searching previously unseen persons across non-overlapping cameras. It is a challenging problem in video surveillance due to the large variations in illumination and viewpoint of personal appearances across cameras (refer Fig. \ref{fig:ReIDChallenge}). 

Most of the existing works on person re-ID generally focus on the design of efficient features and representations \cite{matsukawa2019hierarchical,GOG,LOMO} or distance metric learning \cite{IRS,NK3ML,LOMO,Zheng:nfst}. The features are designed to be robust against the viewpoints and illumination changes of persons. In order to attain better discrimination of the persons, distance metric learning methods are used such that similar class samples come closer and dissimilar class samples get well separated.   

Supervised methods for person re-ID have achieved tremendous improvement in performance. However, for the supervision, obtaining large cross-view annotated data for each of the targeted camera pairs is practically very expensive. Nevertheless, unlabelled data is easily available in abundance and hence methods that can efficiently utilize information from unlabelled data with minimal supervision becomes highly relevant for practical re-ID applications. Therefore, we look at the following problem: Given a few cross-view labelled and a major unlabelled training data (with input features), can we learn a good semi-supervised distance metric for small scale person re-identification, especially when there exists small sample size (SSS) problem, i.e. the number of training samples being very small compared to the feature dimension.

For metric learning methods, the small sample size (SSS) problem can create serious degradation since it creates singularity in the inherent matrices, making it non-invertible \cite{NK3ML,IRS,Zheng:nfst}. In such cases, application of regularization on the matrices or unsupervised dimensionality reduction on the input features makes it sub-optimal due to the loss of discriminative information. The SSS problem becomes more important for small scale semi-supervised case due to the following reason: Though the deeply learned features \cite{ImprDeep,SLSTM, Shi, SCNN,TPC,DGD,Beyond:triplet_loss} perform well with large training data, they struggle to perform with few training samples \cite{NK3ML,matsukawa2019hierarchical, DefenseofColorNames, IRS} and hence hand-engineered features \cite{matsukawa2019hierarchical,GOG,LOMO} are usually employed for small scale person re-identification \cite{matsukawa2019hierarchical,LOMO,GOG}.  However, these hand-engineered features are very high dimensional, typically in ten thousands, much higher than the number of training samples. Furthermore, for a semi-supervised case, the amount of labeled data available for training becomes even lesser compared to the supervised case.

In this paper, we propose a semi-supervised metric learning approach where it first learns a primary cross-view invariant discriminative subspace using the labelled data, by collapsing the same-class samples to singular points in a discriminative nullspace and then maximizing the margin between the classes in a high dimensional kernel space. A secondary discriminative subspace is then learned that maximizes the margin of distinct identities belonging to the unlabelled training data. New pseudo-classes are mined using cross-view affinity learning, which are then used for recursive update of the distance metric to further extract the information in the unlabelled data. Experimental results on four challenging datasets show that our approach outperforms the state-of-the-art methods, with a huge margin.

\section{Related Methods}
\textbf{Supervised methods}:
There exists extensive work on supervised metric learning methods \cite{NK3ML,IRS,Zheng:nfst,LOMO,rPcca,MFML,CRAFT,KISSME}. Using the input features, they learn a  Mahalanobis distance based similarity metric \cite{KISSME,LOMO,MLAPG} or a discriminative space \cite{NK3ML,IRS,rPcca,Zheng:nfst} that well separates the dissimilar class samples and bring same class samples closer. Few metric learning methods \cite{Zheng:nfst, NK3ML, IRS} addressed the small sample size problem in re-ID. Zhang \etal \cite{Zheng:nfst} learned a discriminative nullspace to minimize the within class variance to the least. NK3ML \cite{NK3ML} learned a more discriminative subspace which simultaneously collapses the within classes samples to singular points as well as maximizes the margin between distinct class samples. IRS \cite{IRS} used an identity regression space for embedding each identity using one-hot feature coding. 

Recently, the deep learning based methods  \cite{song2018mask,li2018harmonious, AACN,DGSRW,shen2018person,shen2018end,chen2018improving,chen2017person} have shown impressive performance on large scale labelled data. They generally learn discriminative features and distance metric jointly. However, they have limited performance on small scale training data. \cite{NK3ML,IRS}. \\

\textbf{Semi-supervised methods}: In the literature, there exist few image based semi-supervised methods based on dictionary learning \cite{SSCDL,IterativeLap} as well as metric learning \cite{li2018semi,Zheng:nfst}. SSCDL\cite{SSCDL} jointly learned coupled dictionaries from both labelled and unlabelled images. Kodirov \etal \cite{IterativeLap} used iterative dictionary learning with graph Laplacian regularization. On the other hand, Li \etal \cite{li2018semi} proposed a region metric learning approach with graph based label propagation to estimate positive neighbors from imbalanced unlabelled data. Zhang \etal \cite{Zheng:nfst} used a self-training approach using nullspace based distance metric to learn pseudo-classes from the unlabelled data and addressed the small sample size problem simultaneously. Dictionary learning and metric learning were jointly used in SBAL\cite{SBAL}. It used a Bayesian latent factor based attribute learning framework to determine a dictionary of attributes adaptively.

The most similar work to ours is \cite{Zheng:nfst}. However, unlike our approach, it does not use any margin maximization between distinct class samples as well as cross-view reciprocal constraints for same-class samples to extract pseudo-classes, and has lesser generalization to test data.

\section{Our Approach}

\subsection{Main idea} 
The key idea of our approach is to automatically mine new classes from unlabelled data with the help of available cross-view labelled data. As the labelled data carry information from multiple cameras, they have cross-view identity discrimination information. Hence, we first learn a primary cross-view invariant discriminative space by first collapsing the within class samples to singular points in a discriminative nullspace and then maximizing the margin between the singular points using a maximum margin criterion over a high dimensional kernel space. Next, we obtain within view identity information of the unlabelled data automatically. Then we learn a secondary cross-view invariant distance metric that maximizes the margin between the within view distinct identities of the unlabelled data more efficiently. Over this secondary space, we then use cross-view reciprocal nearest neighbor constraints to discover new possible pseudo-classes. Most confident pseudo-classes are filtered and augmented to the available labelled training data, which is then used to learn a better cross-view discriminative space iteratively.

\begin{figure*}[th!]
\begin{center}
   \includegraphics[width=0.9\linewidth]{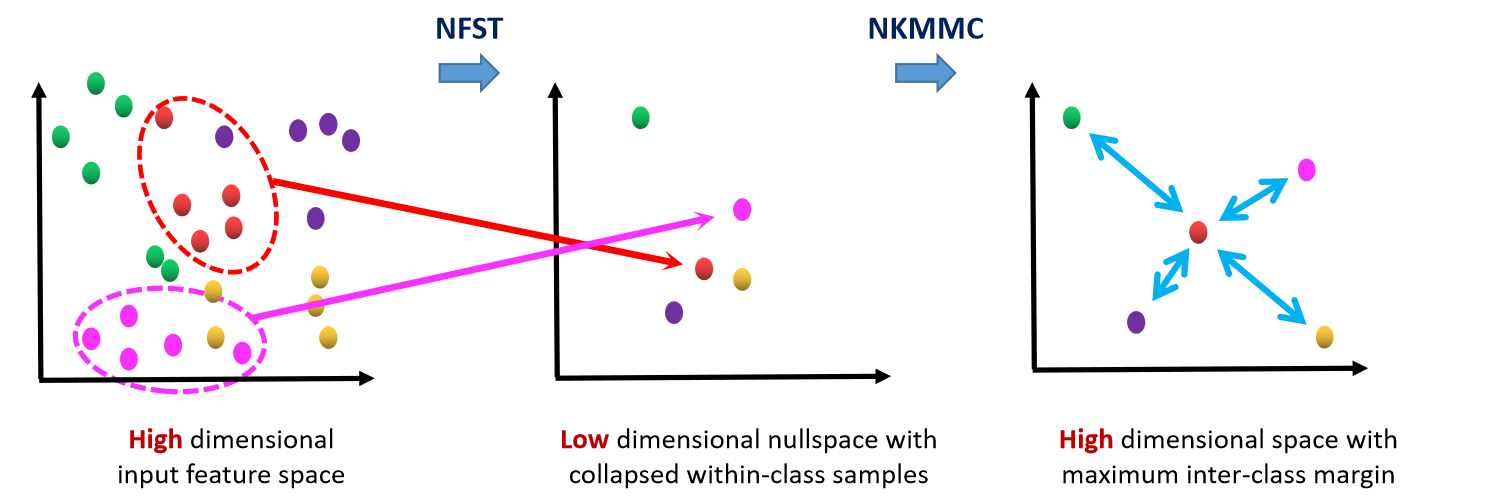}
\end{center}
   \caption{Illustration of Nullspace Kernel Maximum Margin Metric Learning (NK3ML). Each color corresponds to distinct classes. The figure is adapted from \cite{NK3ML}.}
\label{fig:NK3ML}
\end{figure*}

\subsection{Nullspace Kernel Maximum Margin Metric Learning}
We first use the given cross-view labelled samples of the training data to learn a cross-view invariant distance metric based on Nullspace Kernel Maximum Margin Metric learning (NK3ML)\cite{NK3ML}, which is one of the state-of-the-art metric learning methods that address small sample size problem in person re-identification. NK3ML first learns a discriminative nullspace that collapses the samples of the same class into a singular point, thus achieving the least value of zero within class variance. Using the nullspace, NK3ML then learns another discriminative space that maximizes the margin between the dissimilar class samples using a maximum margin criterion. It also uses mapping to very high dimensional space using kernels to obtain maximum inter-class discrimination. NK3ML is illustrated in Fig. \ref{fig:NK3ML}. We briefly explain NK3ML, originally published in \cite{NK3ML}, in the next two subsections for the completeness of the descriptions.

\subsubsection{Null-Foley Sammon Transform}
To collapse the within-class samples to singular points, NK3ML makes use of Null Foley Sammon Transform (NFST)\cite{guo:nfst}. For given $n$ training samples $\mathbf{x} \in \mathbb{R}^d$ belonging to $c$ classes, the objective of NFST is to find discrminants $\mathbf{w}\in \mathbb{R}^d$ that maximize Fisher criterion $\mathcal{J}_F(\mathbf{w})$ with orthonormal constraints.
 \begin{equation}
\begin{aligned}
& \underset{\mathbf{w}_{j}^T \mathbf{w}_{i} = 0,||\mathbf{w}_j|| = 1}   {\text{maximize}} & & \mathcal{J}_F(\mathbf{w}_j) \quad i = 1, \ldots, j-1 \, ,
\label{eqn:OptFST}
\end{aligned}
\end{equation}
where Fisher criterion $\mathcal{J}_F(\mathbf{w})$ is defined as the ratio of between class scatter and within class scatter.
\begin{equation}
\label{eqn:FC}
\mathcal{J}_F(\mathbf{w})  = \dfrac{\mathbf{w}^T \mathbf{S}_b \mathbf{w}}{\mathbf{w}^T \mathbf{S}_w \mathbf{w}} \,.
\end{equation}
$\mathbf{S}_b$ is the between class scatter matrix and $\mathbf{S}_w$ is the within class scatter matrix. 
NFST optimizes the cost function $\mathcal{J}_F(\mathbf{w})$ by forcing the denominator term $\mathbf{w}^T \mathbf{S}_w \mathbf{w}$ to zero while ensuring the numerator term $\mathbf{w}^T \mathbf{S}_b \mathbf{w}$ is positive. 
\begin{align}
\label{eqn:NFSTC1}
\mathbf{w}^T \mathbf{S}_w \mathbf{w} = 0, \,\\
\mathbf{w}^T \mathbf{S}_b \mathbf{w} > 0 \,.
\end{align}
 Thus it attempts to make $\mathcal{J}_F(\mathbf{w})\rightarrow \infty$. The least value of zero  within class variance is attained which results in collapsing the samples of each class to singular points. The total class scatter matrix $\mathbf{S}_t$ is defined as  $\mathbf{S}_t = \mathbf{S}_b + \mathbf{S}_w $. 
If $\mathbf{Z}_t$ and  $\mathbf{Z}_w $  are the nullspaces of $\mathbf{S}_t$ and $\mathbf{S}_w$ respectively, and $\mathbf{Z}_t^\perp$ is the orthogonal compliment of $\mathbf{Z}_t$, then it has been shown that $\mathbf{w} \in (\mathbf{Z}_t^\perp \cap \mathbf{Z}_w)$ \cite{guo:nfst}. The discriminants $\mathbf{w}$ are called Null Projecting Directions (NPDs). When small sample size (SSS) problem occurs, there exists exactly $c-1$ NPDs, where $c$ is the number of classes.

The procedure for obtaining the Null Projecting Directions (NPDs) are as follows. Please refer \cite{NK3ML,guo:nfst} for the detailed theory.\\
1. Take the zero mean data $\mathbf{Y}_t = \{\mathbf{x}_{1}-\mathbf{m},\ldots, \mathbf{x}_{n}-\mathbf{m}\}$, where $\mathbf{m}$ is the mean of the samples.\\
2. Find orthonormal basis for $\mathbf{Y}_t$ to get   $\mathbf{U} = (\theta_1,\ldots, \theta_{n-1})$ using Gram Schmidt orthogonalization procedure.\\
3. Find nullspace of $ \mathbf{U}^T \mathbf{S}_w \mathbf{U}$ to get orthonormal basis vectors $\bm{\beta}$, where each basis vector is of form $\bm{\beta} = (b_1 , \ldots,b_{n-1})$. Since the dimension of the nullspace is $c-1$, there would be $c-1$ solutions for $\bm{\beta}$.\\
4. Each $\bm{\beta}$ corresponds to an NPD $\mathbf{w}$ , which is obtained as 
\begin{equation}
\label{eqn:beta}
\mathbf{w} = b_1 \theta_1 + \ldots + b_{n-1} \theta_{n-1} = \mathbf{U}\bm{\beta}\,.
\end{equation}
The $c-1$ NPDs constitute the discriminative nullspace where the samples of each class collapse to one singular point. Hence we have $c$ singular points in the $c-1$ dimensional discriminative nullspace. 
A general input sample with a high dimensional feature can be mapped to the low dimensional discriminative nullspace using a projection matrix $\mathbf{W}_N \in \mathbb{R}^{d \times (c-1)}$, which is formed by the $c-1$ NPDs as its columns.

\subsubsection{Kernel Maximum Margin Criterion}
Though NFST is optimal in minimizing the within class variance, it fails to maximize the between class variance. As a result, singular points of distinct classes can lie closer in the nullspace. This lack of between class discrimination can result in low generalization for the test data. In order to address this issue, NK3ML uses a \textit{Maximum Margin Criterion} (MMC)\cite{haifeng:mmc} to learn another discriminative space using the nullspace that separates the singular points with maximum margin. In order to attain further separation of the singular points, NK3ML make benefit of kernel based technique that maps the singular points to a very high dimensional space (using an appropriate kernel), such that they are more likely to get well separated with maximum margin.

The Maximum Margin Criterion for $c$ classes is defined as 
\begin{equation}
\label{eqn:MMC1}
\mathcal{J}_M  = \frac{1}{2} \sum\limits_{i=1}^{c} \sum\limits_{j=1}^{c}  \textit{p}_i \textit{p}_j d(C_i,\mathcal{C}_j) \,,
\end{equation}
where the inter-class margin  (or distance) $d(C_i,C_j)$ between the classes $C_i$ and $C_j$ is defined as
\begin{equation}
\label{eqn:MMC2}
d(C_i,\mathcal{C}_j) = d(\mathbf{m}_i,\mathbf{m}_j) - s(C_i) - s(C_j) \,.
\end{equation}
$\mathbf{m}_i$ represents the mean of the class $i$, and $d(\mathbf{m}_i,\mathbf{m}_j)$ is the distance between the class means.  The scatter $s(C_i)$ of class $i$ is estimated as $s(C_i) = tr(\mathbf{S}_i)$, where $\mathbf{S}_i$ is its within class scatter matrix. The margin is solved to get $d(C_i,C_j)  = \textit{tr} \;  (\mathbf{S}_b  -  \mathbf{S}_w)$. 

A set of discriminants $\{\mathbf{v}_k | k=1,\ldots,l\}$ can be than found that maximizes the margin, using the below optimization problem.
\begin{equation}
\begin{aligned}
& \underset{\mathbf{v}_{k}}{\text{maximize}} & &\sum\limits_{k=1}^l \;  \mathbf{v}^T_{k} (\mathbf{S}_b  -  \mathbf{S}_w) \mathbf{v}_{k} \,,\\
\label{eqn:OptMMC}
& \text{subject to}     & & \mathbf{v}^T_{k} \mathbf{v}_{k} = 1 \,, \qquad k=1,\ldots,l \ .
\end{aligned} 
\end{equation}
The optimal solutions are composed of the first $l$ eigenvectors of $\mathbf{S}_b  -  \mathbf{S}_w$.

Instead of using the above Maximum Margin Criterion (MMC) directly, NK3ML utilizes the following \textit{Normalized Kernel Maximum Margin Criterion} (NKMMC) to take benefit of kernel based methods. 
NKMMC learns non-linear discriminants by mapping input sample $\mathbf{y}$ to very high dimensional feature space $\mathcal{F}$ using a non-linear mapping $\phi(\mathbf{y})$, which is implicitly determined by a kernel function  $k(\mathbf{y}_i, \mathbf{y}_j) = \langle \phi(\mathbf{y}_i), \phi(\mathbf{y}_j) \rangle $. Given $n$ training samples, the kernel matrix $\mathbf{K}\in \mathbb{R}^{n \times n}$ can be calculated. Then a discriminant vector $\mathbf{v}_k \in \mathcal{F}$ can be expressed as 
 $\mathbf{v}_k = \sum _{j=1}^{n} \bm{\alpha}_{k}^{(j)} \phi(\mathbf{y}_j)$, where  $\bm{\alpha}_{k}^{(j)}$ is the $j$th component of the expansion coefficient vector $\bm{\alpha}_{k} \in \mathbb{R}^{n}$. Then (\ref{eqn:OptMMC}) can be kernelized to obtain the optimization problem corresponding to NKMMC as:
\begin{equation}
\begin{aligned}
& \underset{\bm{\alpha}_{k}}{\text{maximize}} & & \sum\limits_{k=1}^l \;  \bm{\alpha}^T_{k} (\mathbf{P}  -  \mathbf{Q}) \bm{\alpha}_{k} \,,\\
\label{eqn:KMMCFinal}
& \text{subject to}     & & \bm{\alpha}^T_{k} \mathbf{K} \bm{\alpha}_{k} = 1 \,,
\end{aligned}
\end{equation}
where $\mathbf{Q}:= \sum_{i = 1}^{c} \frac{1}{n} \mathbf{K}_i(\mathbf{I}_{n_i}- \frac{1}{n_i} \mathbf{1}_{n_i}\mathbf{1}_{n_i}^T)\mathbf{K}_i^T$. The matrix  $\mathbf{K}_i \in \mathbb{R}^{n \times n_i}$ corresponding to the $i$th class with $n_i$ samples is defined as $\mathbf{K}_i(p,q):=k(\mathbf{y}_p,\mathbf{y}_q^{(i)})$.  $\mathbf{I}_{n_i}$ is $n_i$ dimensional identity matrix, $\mathbf{1}_{n_i}$ is $n_i$ dimensional column vector of ones, $\mathbf{P}=\sum_{i = 1}^{c} \frac{1}{n_i} (\widetilde{\mathbf{m}}_i-\widetilde{\mathbf{m}})(\widetilde{\mathbf{m}}_i- \widetilde{\mathbf{m}})^T$; $\widetilde{\mathbf{m}} = \frac{1}{n} \sum_{i=1}^{c} n_i \widetilde{\mathbf{m}}_i$ and $\widetilde{\mathbf{m}}_i^{(j)} = \frac{1}{n_i}  \sum_{\mathbf{y} \in C_i}    k(\mathbf{y},\mathbf{y}_j)$. The optimal solutions $\bm{\alpha}_{k}$ are obtained by finding the \textit{generalized} eigenvectors corresponding to the $l$ largest \textit{generalized} eigenvalues of $(\mathbf{P}  -  \mathbf{Q})$ and $\mathbf{K}$. 
In order to obtain maximum margin between the samples, all the eigenvectors with positive eigenvalues are chosen as the discriminants \cite{NK3ML}.

In effect, the NK3ML first maps the input samples from the very high dimensional input feature space to a very low dimensional space, where samples of each class collapse to 
singular points. Then it maps the data from the low dimensional space to a very high dimensional space using appropriate kernels to obtain maximum margin between the singular points. We refer discriminative space of NK3ML, obtained using the labelled training data as the \textit{primary} discriminative space.

\subsection{Pseudo Class Mining}
We next utilize the \textit{unlabelled} data and the primary discriminative space to mine new pseudo-classes in following \textit{five} stages.  

In the \textit{first} stage, we map each of the unlabelled training data $\mathbf{z}\in \mathbb{R}^d $ to the primary space of NK3ML, to obtain $\mathbf{u}\in \mathbb{R}^l $ in two steps: (1) projecting $\mathbf{z}$ onto the NFST nullspace as
\begin{equation}
 \mathbf{y} = \mathbf{W}^T_{N} \mathbf{z}
 \end{equation}
  and then mapping it to each of the discriminant $\mathbf{v}_k$ of NKMMC.
\begin{align}
\mathbf{v}_k^T \phi(\mathbf{y}) &= \Big( \sum\limits_{j = 1}^{n} \bm{\alpha}_{k}^{(j)} \phi(\mathbf{y}_j)\Big)^T \phi(\widetilde{\mathbf{y}})\\
&= \sum\limits_{j = 1}^{n} \bm{\alpha}_{k}^{(j)}  k(\mathbf{y}_j,\widetilde{\mathbf{y}})
\end{align}

\noindent \textbf{Within-view identity labeling}: 
 In the second stage, we propose to use the original input images/videos of the unlabelled samples from each of the  given camera to automatically identify the images that belong to distinct persons.  Though the true identities of the images are not known, standard intra-camera tracking models with spatial and temporal constraints of the cameras can be used to extract person tracklets and annotate every tracklet with a distinct stamp \cite{TAUDL}. Thus within view identity information can be extracted to group the samples to distinct classes. Note that the cross-view identity association is still not known and need to be estimated.\\

\subsubsection{Anchor-view identity discrimination} 
In the \textit{third} stage, we choose  the camera with the maximum number of distinct person appearances as an \textit{anchor camera}. Then we learn a \textit{secondary} discriminative space that better maximizes the distance between the within view identities of the anchor camera. It should be noted that though the primary space is already learned using the initial labelled training data $\mathbf{x}$ to discriminate distinct identities, it could not use the data $\mathbf{z}$ for its training. Hence certain distinct identities from data $\mathbf{z}$ may still be not well separated.  
Hence we learn the secondary discriminative space that is more suitable in separating the distinct identities in $\mathbf{z}$ coming from the anchor camera.

To that end, we use the same Normalized Kernel Maximum Margin Criterion (NKMMC) in (\ref{eqn:KMMCFinal}), however, by using the anchor camera samples $\mathbf{u}^A$ lying in the primary discriminative space as the training data.
As seen previously, the Normalized Kernel Maximum Margin Criterion (NKMMC) can ensure separation of distinct training classes with maximum margin.  
Given $\tilde{n}$ samples belonging to $\widetilde{c}$ classes from the anchor camera, we map the its data samples $\mathbf{u}^A$ to a secondary kernel space $\widetilde{\mathcal{F}}$, using a secondary non-linear mapping function $\widetilde{\phi}(\mathbf{u}^A)$. Using this space, it learns $f$ discriminant vectors $\{\widetilde{\mathbf{v}}_k\}_{k=1}^f$ that maximize $\mathcal{J}_M$. Since each discriminant $\widetilde{\mathbf{v}}_k$ lies in the span of the mapped samples, we have $\widetilde{\mathbf{v}}_k = \sum_{j=1}^{{\tilde{n}}} \epsilon_{k}^{(j)} \widetilde{\phi}(\mathbf{u}^A_j)$
where $\epsilon_{k}^{(j)}$ is the \textit{j}th element of the expansion vector $\bm{\epsilon}_k \in \mathcal{R}^{\tilde{n}}$. Let $\widetilde{\mathbf{K}}\in \mathcal{R}^{\tilde{n} \times \tilde{n}}$ be the kernel matrix computed using the secondary kernel function $\widetilde{k}(\mathbf{u}^A_i,\mathbf{u}^A_j)$. Then the optimization problem based on NKMMC to learn the optimal discriminants $\widetilde{\mathbf{v}}_k$ is
\begin{equation}
\begin{aligned}
& \underset{\bm{\epsilon}_{k}}{\text{maximize}} & & \sum\limits_{k=1}^f \;  \bm{\epsilon}^T_{k} (\widetilde{\mathbf{P}}  -  \widetilde{\mathbf{Q}}) \bm{\epsilon}_{k}\\
\label{eqn:CKMMCFinal}
& \text{subject to}     & & \bm{\epsilon}^T_{k}\widetilde{\mathbf{K}} \bm{\epsilon}_{k} = 1
\end{aligned}
\end{equation}
where $\widetilde{\mathbf{Q}}=\sum_{i = 1}^{\widetilde{c}} \frac{1}{\widetilde{n}} \widetilde{\mathbf{K}}_i(\mathbf{I}_{\widetilde{n}_i}-\frac{1}{\widetilde{n}_i} \mathbf{1}_{\widetilde{n}_i}\mathbf{1}_{\widetilde{n}_i}^T)\widetilde{\mathbf{K}}_i^T$, 
 $\widetilde{\mathbf{P}}=\sum_{i = 1}^{\widetilde{n}} \frac{1}{\widetilde{c}_i} (\widetilde{\mathbf{m}}_i-\widetilde{\mathbf{m}})(\widetilde{\mathbf{m}}_i- \widetilde{\mathbf{m}})^T$, $\widetilde{\mathbf{m}} = \frac{1}{\widetilde{c}} \sum_{i=1}^{\widetilde{c}} \widetilde{c}_i \widetilde{\mathbf{m}}_i$, $ \widetilde{\mathbf{m}}_i^{(j)} = \frac{1}{\widetilde{c}_i}  \sum_{\mathbf{u}^A \in C_i}    \widetilde{k}(\mathbf{u}^A,\mathbf{u}^A_j)$.
Similar to (\ref{eqn:KMMCFinal}), the optimal solutions are composed of the generalized eigen vectors corresponding to the leading $f$ generalized eigenvalues of $\widetilde{\mathbf{P}}-\widetilde{\mathbf{Q}}$ and $\widetilde{\mathbf{K}}$.
We use all the eigenvectors with positive eigenvalues to attain a good inter-class margin.

\begin{figure}[t]
\begin{center}
   \includegraphics[width=0.95\linewidth]{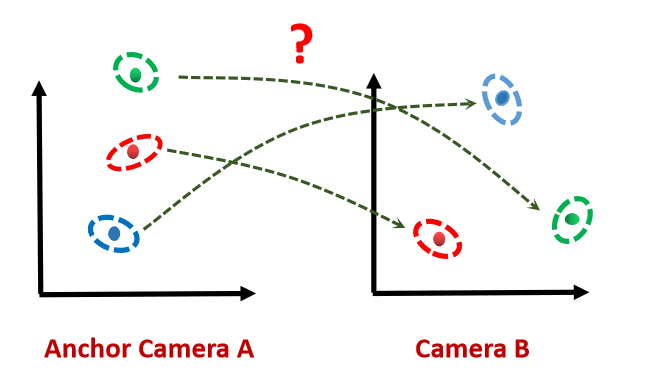}
\end{center}
   \caption{Cross-view affinity learning of the identities from the anchor camera with respect to another camera (each color represents distinct identities). }
\label{fig:AffinityLearn}
\end{figure}

\begin{figure*}[ht!]
\centering
\begin{subfigure}{0.5\linewidth}
  \centering
  \includegraphics[width=0.5\linewidth]{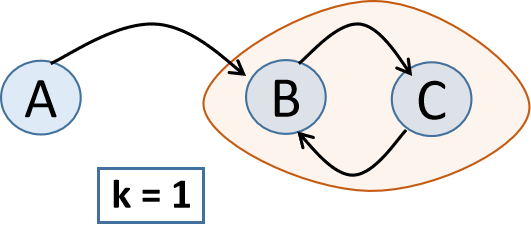}
  \caption{}
  \label{fig:sub1}
\end{subfigure}%
\begin{subfigure}{.5\linewidth}
  \centering
  \includegraphics[width=0.8\linewidth]{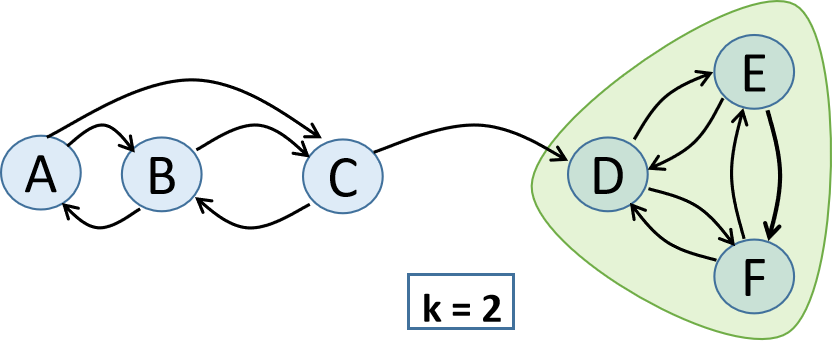}
  \caption{}
  \label{fig:sub2}
\end{subfigure}
\caption{Illustration of k-reciprocal nearest neighbor constraint: The unidirectional k-nearest neighbors of each node is shown using the arrows. (a) For k=1, the node A does not have any reciprocal neighbors as its unidirectional neighbor B does not have A as the  unidirectional neighbor. However, nodes B and C are k-reciprocal neighbors as each of them is a k-nearest neighbor of the other.  (b) For k=2, 
nodes A and C are k-reciprocal neighbors of B, while A and C are not k-reciprocal neighbors of each other. However, nodes C, D and E are all k-reciprocal nearest neighbors of each other. Figure 4(b) is adapted from \cite{helloneighbor}.}
\label{fig:kreciprocal}
\end{figure*}

We then map all the unlabelled samples $\mathbf{u}$ from all the cameras to this secondary space. A general sample $\mathbf{u}$ can be projected onto the discriminant vector $\widetilde{\mathbf{v}}_k$ as 
\begin{align}
\widetilde{\mathbf{v}}_k^T \widetilde{\phi}(\mathbf{u}) &= \Big(\sum_{j=1}^{\tilde{n}}  \epsilon_k^{(j)} \widetilde{\phi}(\mathbf{u}_j^A) \Big )^T \widetilde{\phi}(\mathbf{u}) 
= \sum\limits_{j = 1}^{\tilde{n}} \bm{\epsilon}_{k}^{(j)} \widetilde{k}(\mathbf{u}_j^A,\mathbf{u}).
\end{align}
All the distinct identity samples coming from the anchor camera are now guaranteed to lie well separated with maximum margin in the secondary space. It should be noted that for both NK3ML and anchor-view identity discrimination, we used mapping to kernel space. Hence there occur two levels of subsequent high dimensional kernel mappings in our approach, which helps in improving the separation of the underlying identities with maximum margin.\\

\subsubsection{Cross-view affinity learning} 
As the secondary space is learned over the view invariant primary space, the secondary space has two important properties: (1) anchor-view identity discrimination property, as the secondary space is better trained with respect to the unlabelled training data identities seen in the anchor camera;  (2) view invariant property, which naturally occurs as the secondary metric is learned over the view invariant primary space.  As a consequence of the two properties existing together in the secondary space, the cross-view unlabelled samples from other cameras tend to lie closer to their corresponding true identities from the anchor camera, as well as all the distinct identity samples get well separated with maximum margin. This results in localization of the cross-view same class samples and hence better discrimination of the cross-view data is achieved in the secondary space. 

In the \textit{fourth} stage, we utilize this secondary space to perform cross-view affinity learning for each of the identities of the anchor camera against all the identities in other cameras, as illustrated in Fig. \ref{fig:AffinityLearn}. We then mine new pseudo-classes across views by using  \textit{cross-view k-reciprocal-nearest-neighbor constraint} over the affinities.\\

\noindent \textbf{Reciprocal-nearest-neighbor constraint}: One way to mine new pseudo-classes in the secondary space is to retrieve nearest neighbors across views and perform k-nearest neighbor clustering for each of the identities of the anchor camera against the other views. Though we have localized the identities in the secondary space, however, due to the large variations in illumination and viewpoint of persons across views, instead of the true matches, irrelevant objects may exist among the k-nearest neighbors, which can degrade the mined pseudo-classes. In order to avoid such degradation and to achieve  a higher and robust performance, we propose to use k-reciprocal-nearest-neighbor constraint \cite{CDM,helloneighbor},   across views, for mining new classes. It helps in finding cross-view samples which are very likely to be related to the queries, by measuring mutual similarity.

The k-nearest neighbors of a query sample $q$ is defined as
\begin{equation}
N_k(q) = \{ g_1, g_2,\ldots,g_k\}, \;\;|N_k(q)| = k
\end{equation}
where $|X|$ represents the cardinality of the set $X$, and $g_1$, $g_2$,$\ldots$,$g_k$ are the top-k search retrieval results for $q$. Then k-reciprocal nearest neighbors of $q$ are defined as 
\begin{equation}
R_k(q) = \{ g_i\,|\, (g_i \in N_k(q)) \wedge (q \in N_k(g_i))\}.
\end{equation}
The concept of k-reciprocal nearest neighbors is illustrated in Fig. \ref{fig:kreciprocal}. Two samples are k-reciprocal nearest neighbors, if for each of the samples as queries, the other sample exists in the former's top k-retrieved closest neighbors. It is a stronger constraint of closeness compared to the conventional k-nearest neighbor relationship. The k-reciprocal nearest neighbor constraint considers local densities around the queries and their retrieved matches. Hence it can deal with uneven distribution of samples in the space, thus becoming more robust to false retrieval results and outliers.

For cross-view affinity learning for the anchor camera, we use the \textit{cross-view} k-reciprocal-nearest-neighbor constraint for searching the most likely matches. The cross-view k-reciprocal-nearest-neighbors for  any identity $q^A$ of the anchor camera with respect to another camera $B$ is defined as
\begin{equation}
R^*_k(q^A) = \{ g_i| \;\; \ g_i  \in R_k(q^A), \;R_k(q^A) \in C_B \},
\end{equation}
where $C_B$ is the set of all identities in camera $B$.

\subsubsection{Recursive metric learning}
\begin{figure}[t!]
\begin{center}
   \includegraphics[width=0.7\linewidth]{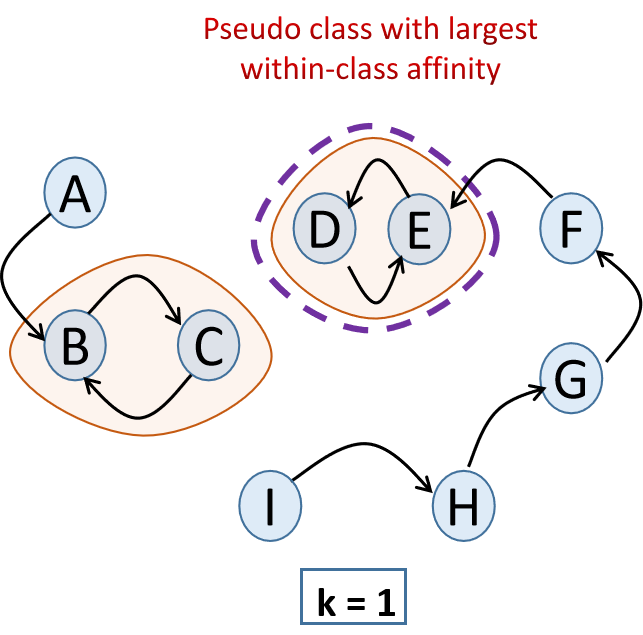}
\end{center}
   \caption{Illustration of the pseudo-class mining: At each iteration, the pseudo-classes with largest intra-class affinities are given new labels and augmented to the labelled training data.}
\label{fig:PseudoClasses}
\end{figure}

In the \textit{fifth} stage, the newly mined cross-view pseudo-classes are ranked according to the increasing order of their within class affinities (refer Fig. \ref{fig:PseudoClasses}). The pseudo-classes with affinities above the upper quartile are filtered and augmented to the existing labelled training data. Next, we learn a new  primary discriminative space using NK3ML with the new set of labelled training classes. All the above steps are repeated recursively till there are no more cross-view pseudo class available in the unlabelled training data that satisfy the k-reciprocal nearest neighbor constraint. 
 
The strategy of selecting a fraction of the top ranked pseudo-classes in each iteration is to ensure that the pseudo-classes in the initial iterations are the best, which in turn help in mining more reliable pseudo-classes in the subsequent iterations. 
This also ensures that model drifting due to errors in the mined pseudo-classes is better avoided, which is of utmost importance in self-training \cite{SSSurvey} based methods.

The proposed method thus utilizes the information from both labelled and unlabelled data to learn a discriminative distance metric effectively. The approach ensures that within class variance is minimized to the least and the distinct identity samples get well separated with maximum margin to get better generalization for test data. Note that both NK3ML and NKMMC that is used for learning the primary and secondary discriminative spaces have closed form solutions, which does not need  regularization or unsupervised dimensionality reduction. Hence our proposed method efficiently handles the small sample size (SSS) problem in person re-identification.

\begin{figure*}[ht!]
\centering
  \begin{subfigure}{.24\linewidth}
  \centering
  \includegraphics[width=0.8\linewidth]{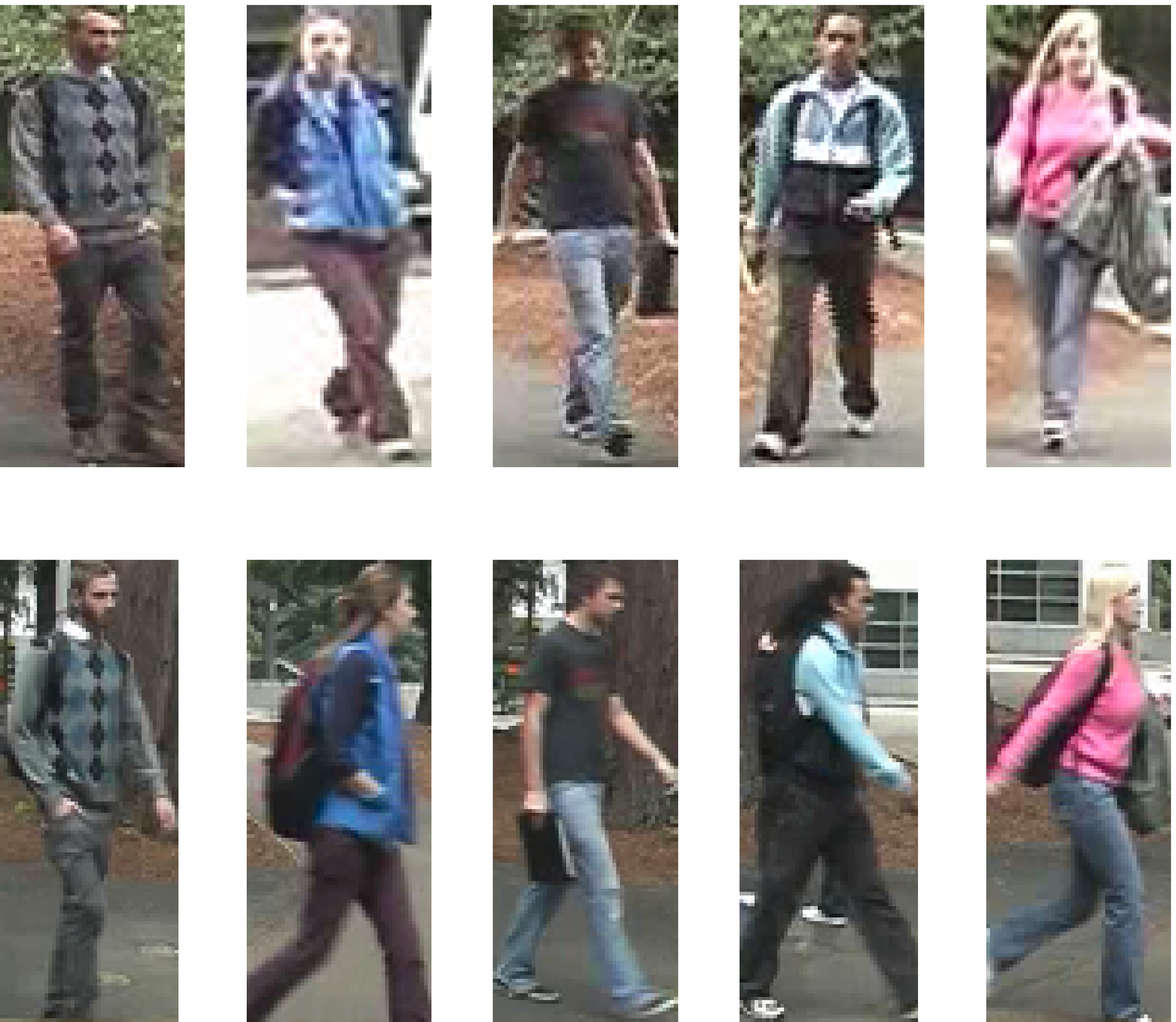}
  \caption{}
  \label{fig:sub2}
\end{subfigure}
\begin{subfigure}{.24\linewidth}
  \centering
  \includegraphics[width=0.8\linewidth]{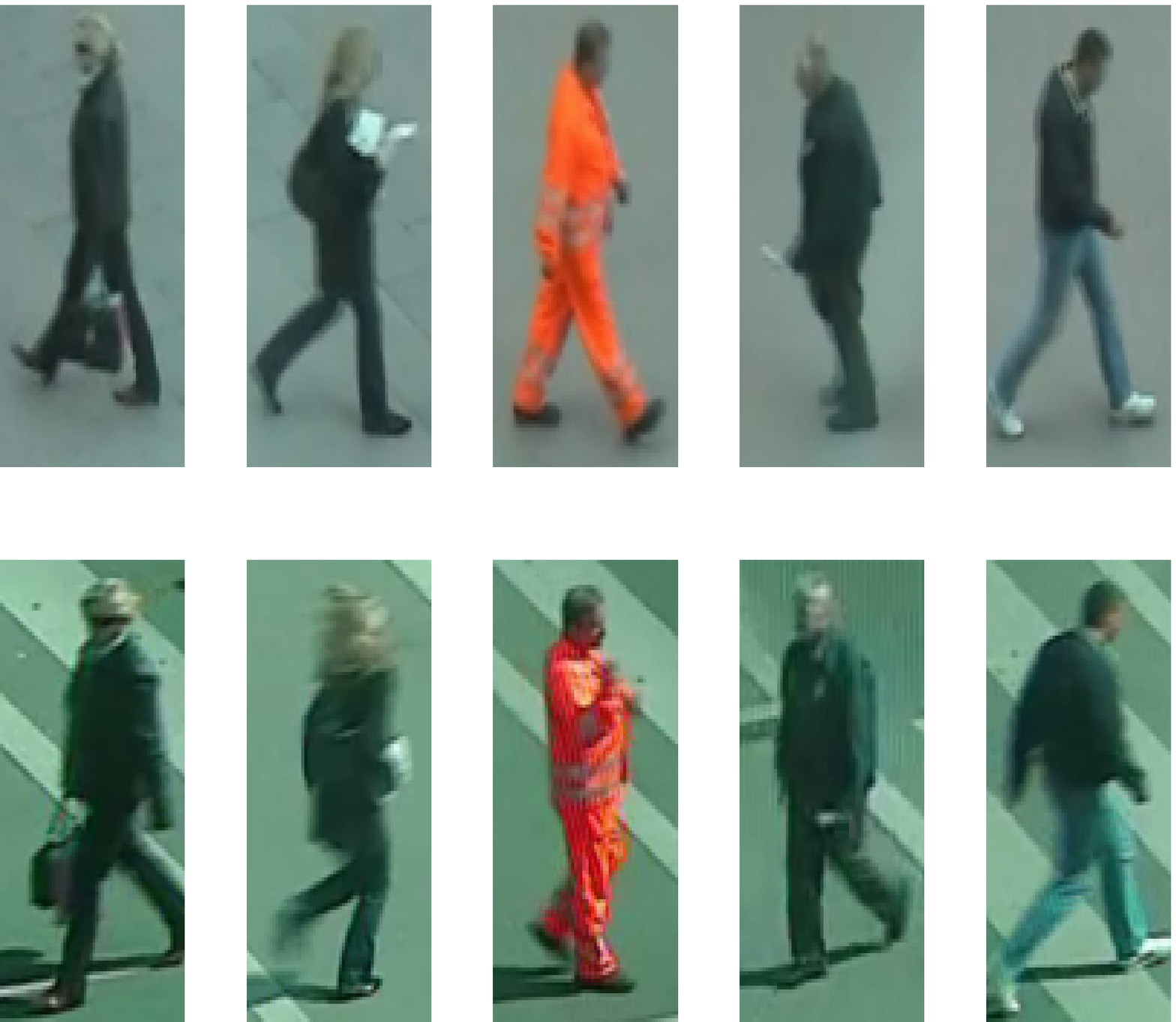}
  \caption{}
  \label{fig:sub2}
  \end{subfigure}
\begin{subfigure}{0.24\linewidth}
  \centering
  \includegraphics[width=0.8\linewidth]{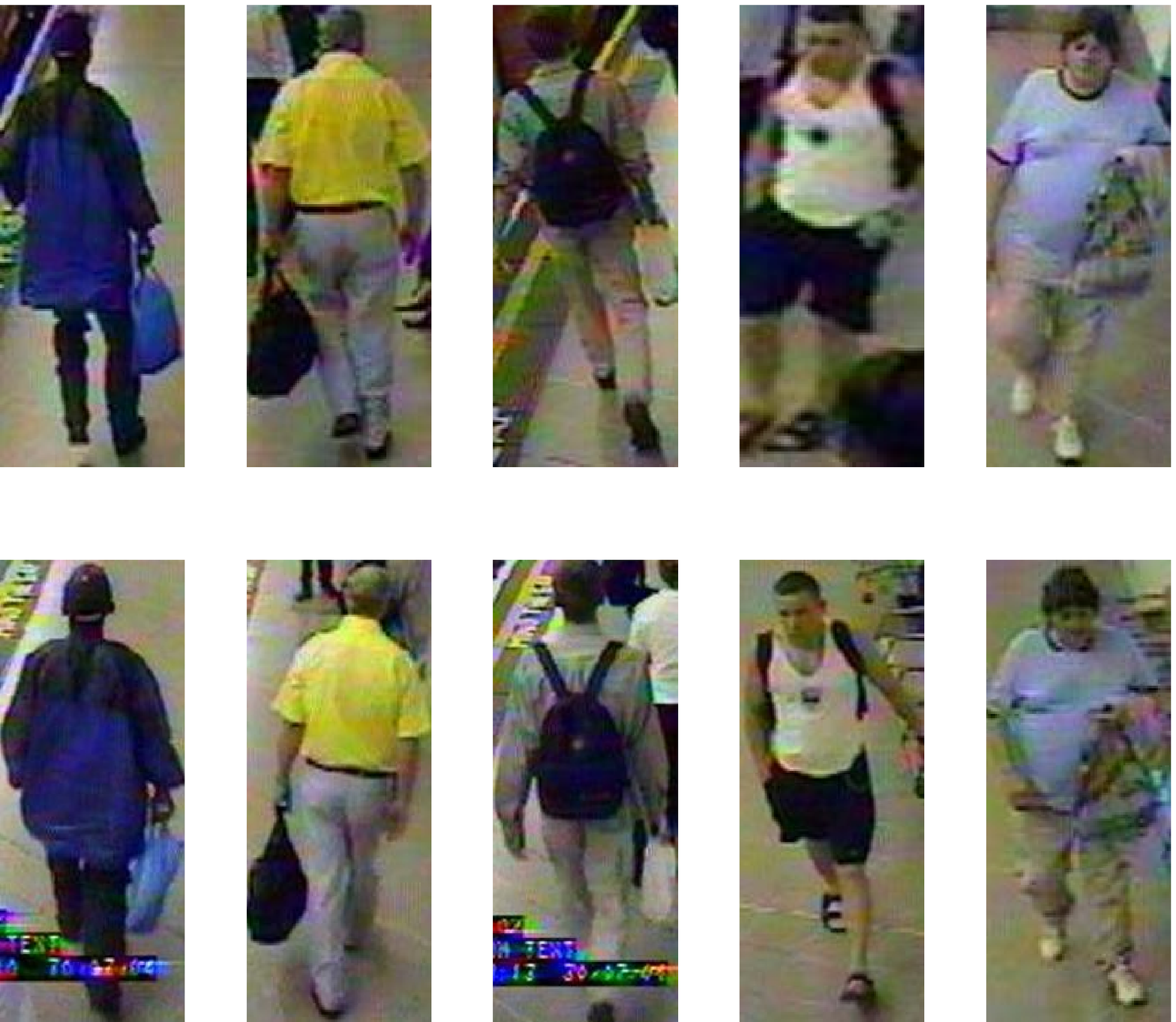}
  \caption{}
  \label{fig:sub1}
\end{subfigure}%
  \begin{subfigure}{.24\linewidth}
  \centering
  \includegraphics[width=0.8\linewidth]{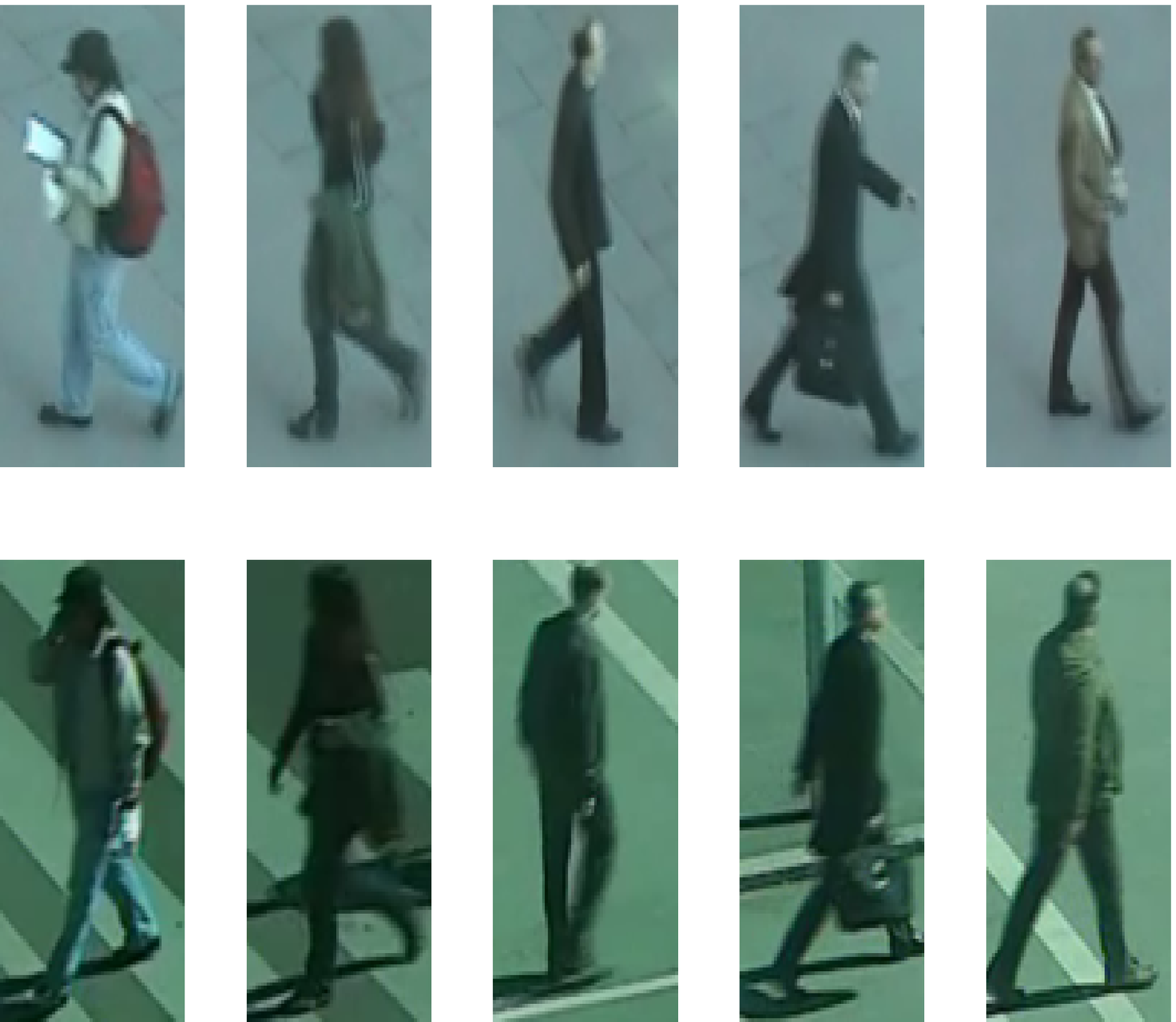}
  \caption{}
  \label{fig:sub2}
\end{subfigure}

\caption{Sample images from (a) VIPeR, (b) PRID2011, (c) GRID and (d) PRID450S datasets.}
\label{fig:datasets}
\end{figure*}

\section{Experiments}
\noindent\textbf{Datasets and Evaluation Protocol}: 
We use four small scale datasets including VIPeR \cite{ELF}, PRID450S \cite{PRID450S}, GRID \cite{GRID1} and PRID2011 \cite{prid2011} for our experiments. They respectively contain 632, 450, 250 and 200 persons, captured from two disjoint cameras, with one image per person in each camera. Following the standard protocol \cite{Zheng:nfst,IterativeLap,SSCDL,SBAL}, the dataset identities are divided equally into two halves; one forming the training set and the other forming the test set. A random one third of the training set is labelled and the rest is unlabelled. 
The test images are divided into half forming the query(probe) set and the gallery set. Images from one camera form the query set and the rest form the gallery. The gallery of GRID and PRID2011 datasets have an additional 775 and 549 person images, respectively, which are different from their query set identities.  Each image of the query set is matched against the gallery. The distance scores are ranked and sorted. Rank-N matching performance is then calculated as the percentage of true match being present in the first N search results. The above procedure is repeated 10 times and the average performance is reported.

\noindent\textbf{Features}: In our method, we use standard feature descriptors including  LOMO\cite{LOMO}, GOG\cite{GOG} and WHOS\cite{LisantiPAMI14}. They are of dimensions 26960, 27622 and 2960 respectively. In addition, we use a new feature descriptor named $\text{LOMO}^{\dagger}$, which is the LOMO feature obtained without using Retinex\cite{LOMO} transformation of the images, in order to take advantage of color diversity, similar to GOG. 
Re-ID datasets have large variations in illumination and background. Hence we use specific features for each dataset, to better capture their intrinsic characteristics. For VIPeR dataset, we use concatenated LOMO and WHOS features, while for other datasets, we use concatenated LOMO, $\text{LOMO}^{\dagger}$ and GOG features. Naturally, small sample size problem exists due to the large dimension of the features.

\noindent\textbf{Parameters}: We use RBF kernel for both NK3ML, as well as for NKMMC in anchor-view identity discrimination. The width of the RBF kernel  is automatically set using a mean function of the training samples, similar to \cite{Zheng:nfst} and \cite{rPcca}. We set k=1 for cross-view reciprocal nearest neighbor constraint.

\begin{table}[t]
\begin{center}
\small
\resizebox{0.98\columnwidth}{!}{%
\begin{tabular}{lcccc}
\hline
Methods & Rank1 & Rank5 &  Rank10 & Rank20 \\
\hline \hline
KISSME\cite{KISSME} &  18.50 & 43.70 & 57.90 & 74.5\\
kLFDA\cite{rPcca} &27.50 & 56.00 & 69.60 & 82.60\\
kCCA\cite{LisantiDSC} & 24.60 & 56.20 &71.70 & 85.60 \\
XQDA\cite{LOMO}& 27.63 & 55.76 & 68.67& 81.90 \\
MLAPG\cite{MLAPG} &25.82 & 48.54 & 59.24 & 70.22 \\
kNFST\cite{Zheng:nfst} & 26.46 &  54.72 & 67.66 & 81.30 \\
\hline
SSCDL\cite{SSCDL}& 25.60  & 53.70 & 68.20 & 83.60\\
IterativeLap\cite{IterativeLap}  &30.06 & 46.84 & 56.74 &67.31\\
semi-kNFST\cite{Zheng:nfst}& 31.68 &59.40 & 72.78&84.91 \\
SBAL \cite{SBAL} & 33.60 &-&- &- \\
\hline
Ours&  64.75 & 88.83 & 94.46& 97.78\\
\hline
\end{tabular}
}
\end{center}
\caption{Performance comparison with state-of-the-art results on VIPeR dataset.}  
\label{table:VIPeR}
\end{table}

\begin{table}[t]
\begin{center}
\small
\resizebox{0.98\columnwidth}{!}{%
\begin{tabular}{lcccc}
\hline
Methods & Rank1 & Rank5 &  Rank10 & Rank20 \\
\hline \hline
KISSME\cite{KISSME} &  5.10& 15.20 & 24.10 & 40.10\\
kLFDA\cite{rPcca} &14.10 & 33.70 & 44.00 & 56.20\\
kCCA\cite{LisantiDSC} & 5.30&15.70 & 25.80 &37.00 \\
XQDA\cite{LOMO} & 13.50 & 30.60 & 40.20 & 52.50 \\
kNFST\cite{Zheng:nfst} & 12.40 & 28.40 & 37.50 & 49.40 \\
\hline
IterativeLap\cite{IterativeLap}  & 22.10 & 45.30 & 56.50 & 66.30 \\
semi-kNFST\cite{Zheng:nfst}& 24.70 & 46.80 & 58.20 & 68.20\\
SBAL \cite{SBAL} & 24.40 &-&- &- \\
\hline
Ours& 29.90 & 53.10 & 61.60 & 72.20\\
\hline
\end{tabular}
}
\end{center}
\caption{Performance comparison with state-of-the-art results on PRID2011 dataset.}
\label{table:PRID2011}
\end{table}

\begin{table}[t]
\begin{center}
\small
\resizebox{0.98\columnwidth}{!}{%
\begin{tabular}{lcccc}
\hline
Methods & Rank1 & Rank5 &  Rank10 & Rank20 \\
\hline \hline
XQDA\cite{LOMO} &  9.92 & 22.08 & 29.92 & 39.92  \\
kNFST\cite{Zheng:nfst} &   9.60 &  20.08 & 27.60 &37.60 \\
IterativeLap\cite{IterativeLap}  & 10.96 & 28.64 & 37.92 & 48.24 \\
\hline
Ours & 23.36 & 39.76 & 52.32 & 63.84\\
\hline
\end{tabular}
}
\end{center}
\caption{Performance comparison with state-of-the-art results on GRID dataset.  }
\label{table:GRID}
\end{table}

\begin{table}[t]
\begin{center}
\small
\resizebox{0.98\columnwidth}{!}{%
\begin{tabular}{lcccc}
\hline
Methods & Rank1 & Rank5 &  Rank10 & Rank20 \\
\hline \hline
IterativeLap\cite{IterativeLap}  & 24.31 & 42.13 & 51.78 & 61.82 \\
XQDA\cite{LOMO} & 45.07 & 71.64 & 81.78   &90.13  \\
MLAPG\cite{MLAPG} & 39.91 & 67.47 & 78.62 & 88.62\\
kNFST\cite{Zheng:nfst} & 41.87 & 70.04 & 81.07 & 90.80  \\
\hline
Ours & 63.78 & 86.49 & 92.53 & 96.80 \\
\hline
\end{tabular}
}
\end{center}
\caption{Performance comparison with state-of-the-art results on PRID450S dataset.}
\label{table:PRID450S}
\end{table}

\begin{table}[t]
\small 
\begin{center}
\begin{tabular}{l|c|c|c|c}
\hline
Methods & VIPeR & PRID2011& GRID & PRID450S\\ 
\hline\hline
IterativeLap\cite{IterativeLap} & 1680 & 975	& 1214	 & 871	\\
Ours & 48	& 21	& 33	&58\\
\hline
\end{tabular}
\end{center}
\caption{Execution time comparison (in seconds) for various datasets.}
\label{table:executionTime}
\end{table}

\subsection{Comparative Results}

The experimental results of our method evaluated on datasets VIPeR, PRID2011, GRID and PRID450S are provided in Table \ref{table:VIPeR}, \ref{table:PRID2011}, \ref{table:GRID} and \ref{table:PRID450S}, respectively. We compare our results with state-of-the-art semi-supervised methods including  SBAL \cite{SBAL}, semi-kNFST\cite{Zheng:nfst}, IterativeLap \cite{IterativeLap} and SSCDL\cite{SSCDL}. RegionMetric \cite{li2018semi} uses pre-training using other datasets and hence is not directly comparable with our method. We also compare with supervised methods including KISSME\cite{KISSME}, kLFDA\cite{rPcca}, kCCA\cite{LisantiDSC}, XQDA\cite{LOMO},
 and kNFST\cite{Zheng:nfst}, whose results are provided in separate rows. These methods are trained using only the labelled training data. The results of KISSME, kLFDA, and kCCA are obtained from \cite{IterativeLap}. For all the datasets, our approach outperforms all the compared methods, with high margin, at all ranks. Especially for VIPeR dataset, we attain an improvement of more than 30\% in rank-1 accuracy, compared to the second best method SBAL\cite{SBAL}. Similarly, a rank-1 improvement margin of more than 5\%, 12\%, and 22\% is attained for PRID2011, GRID and PRID450S datasets, respectively. It is worth noting that SBAL\cite{SBAL} uses deep learning based features and our method still outperforms it with a good margin. This emphasizes the limitation of deep learning methods for semi-supervised person re-identification with minimal training data. Our method is able to utilize the information from both the available labelled and unlabelled data, while efficiently handling the small sample size problem in person re-identification.

\subsection{Execution Time}
We evaluate the total run time of our method in Table \ref{table:executionTime}. IterativeLap\cite{IterativeLap} is the only semi-supervised method whose source code we could find publicly available and hence we use the same for comparison. Both the methods are implemented in MATLAB and evaluated in a PC with 3.4GHz CPU and 32GB memory. It can be seen that our method has low time complexity and is more suitable for practical implementation.

\section{Conclusion}
In this paper, we proposed a semi-supervised approach for person re-identification that learns an efficient cross-view invariant distance metric for handling small training sample size problem. Using a few labelled training samples, the method first learns a primary cross-view discriminative space by collapsing the training classes into singular points and then maximizing the margin between the singular points. Later a secondary discriminative space is learned that separates the distinct identities in the unlabelled data. Cross-view affinity learning with k-reciprocal nearest neighbor constraints is used to mine new pseudo-classes, which are augmented to the labelled training data to learn a better cross-view discriminative distance metric recursively. The  method does not need regularization or unsupervised dimensionality reduction and handles small sample size problem efficiently. Experiments on challenging datasets confirm the superiority of our approach compared to the state-of-the-art methods.\\

\noindent\textbf{Acknowledgment}: This research work is supported under Visvesvaraya PhD Scheme for Electronics and IT, by Ministry of Electronics and Information Technology (MeitY), Government of India.

{\small
\bibliographystyle{ieee_fullname}
\bibliography{egbib_ICCVW}
}

\end{document}